\documentclass[letterpaper, 10 pt, conference]{ieeeconf}
\IEEEoverridecommandlockouts    
\usepackage{graphics}           
\usepackage{times}              
\usepackage{amsmath}            
\usepackage{amssymb}            
\usepackage{graphicx}
\usepackage{algorithm}
\usepackage[noend]{algpseudocode}
\usepackage{booktabs}
\usepackage{balance}
\usepackage{color}
\usepackage[nocompress]{cite} 
\definecolor{instructioncolor}{rgb}{.5,.5,.5}
\usepackage{multirow} 
\usepackage{url}

\usepackage[font=small]{caption}
\usepackage{svg}

\def\eqref#1{Eq.~(\ref{#1})}

\makeatletter
\usepackage{xspace}
\DeclareRobustCommand\onedot{\futurelet\@let@token\@onedot}
\def\@onedot{\ifx\@let@token.\else.\null\fi\xspace}

\makeatother

\usepackage{array}
\newcolumntype{L}[1]{>{\raggedright\let\newline\\\arraybackslash\hspace{0pt}}m{#1}}
\newcolumntype{C}[1]{>{\centering\let\newline\\\arraybackslash\hspace{0pt}}m{#1}}
\newcolumntype{R}[1]{>{\raggedleft\let\newline\\\arraybackslash\hspace{0pt}}m{#1}}

\usepackage{adjustbox}

\title{\LARGE \bf Semantic Landmark Particle Filter for Robot Localisation in Vineyards}
\author{Rajitha de Silva$^{1}$ \and Jonathan Cox$^{1}$ \and James R. Heselden$^{1}$ \and Marija Popovi\'{c}$^{2}$ \and Cesar Cadena$^{3}$ \and Riccardo Polvara$^{1}$
  \thanks{$^{1}$Rajitha de Silva, Jonathan Cox, James R. Heselden and Riccardo Polvara are with Lincoln Centre for Autonomous Systems (L-CAS), University of Lincoln, UK. $^{2}$Marija Popovi\'{c} is with MAVLab, TU Delft, Netherlands. $^{3}$Cesar Cadena is with Robotics Systems Lab, ETH Zurich, Switzerland. (correspondence author: Rajitha de Silva {\tt\small $^{1}$odesilva@lincoln.ac.uk})}%
  \thanks{This work was supported by Engineering and Physical Sciences Research Council (EPSRC), UK Project: ``GAIA: Ground-Aerial maps Integration for increased Autonomy outdoors" (EPSRC Reference: EP/Y003438/1).}
}
\begin{document}
\maketitle
\thispagestyle{empty}
\pagestyle{empty}

\begin{abstract}
  %


  Reliable localisation in vineyards is hindered by row-level perceptual aliasing: parallel crop rows produce nearly identical LiDAR observations, causing geometry-only and vision-based SLAM systems to converge towards incorrect corridors, particularly during headland transitions. We present a Semantic Landmark Particle Filter (SLPF) that integrates trunk and pole landmark detections with 2D LiDAR within a probabilistic localisation framework. Detected trunks are converted into \emph{semantic walls}, forming structural row boundaries embedded in the measurement model to improve discrimination between adjacent rows. GNSS is incorporated as a lightweight prior that stabilises localisation when semantic observations are sparse.

  Field experiments in a 10-row vineyard demonstrate consistent improvements over geometry-only (AMCL), vision-based (RTAB-Map), and GNSS baselines. Compared to AMCL, SLPF reduces Absolute Pose Error by 22\% and 65\% across two traversal directions; relative to a NoisyGNSS baseline, APE decreases by 65\% and 61\%. Row correctness improves from 0.67 to 0.73, while mean cross-track error decreases from 1.40\,m to 1.26\,m. These results show that embedding row-level structural semantics within the measurement model enables robust localisation in highly repetitive outdoor agricultural environments.



\end{abstract}

\section{Introduction}
\label{sec:intro}

Accurate localisation is essential for long-term autonomy in vineyards and orchards, where robots perform monitoring, spraying, and yield estimation. However, vineyards consist of highly repetitive parallel rows that induce severe perceptual aliasing, making adjacent corridors geometrically indistinguishable for geometry-only localisation systems. Seasonal vegetation changes further reduce robustness by introducing transient structure while leaving only trunks and support poles spatially stable.

\begin{figure}[t]
  \centering
  \includegraphics[width=0.99\linewidth]{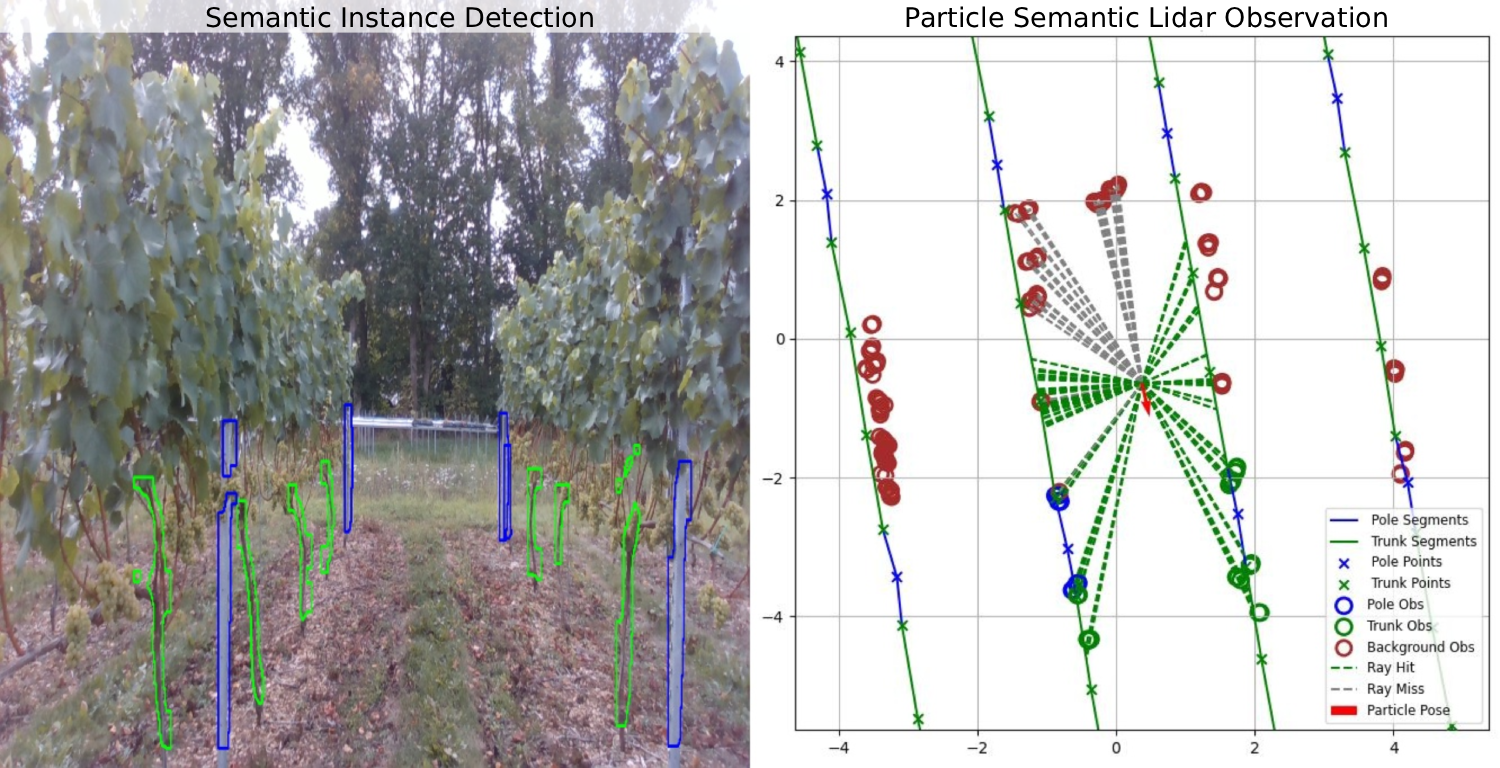}
  \caption{Vineyard semantics detection (left) projected onto the landmark map (right) for likelihood calculation. Green: vine trunks, Blue: support poles.}
  \label{fig:motivation}
\end{figure}

LiDAR-based localisation remains widely used due to its robustness to illumination changes, yet repetitive crop geometry leads to persistent wrong-row hypotheses that standard likelihood models struggle to reject \cite{aguiar2022localization, nehme2021lidar}. As illustrated in Fig.~\ref{fig:motivation}, adjacent vineyard rows generate nearly identical LiDAR signatures under geometry-only models. Recent work has explored LiDAR-vision fusion to mitigate these limitations. Vision-based trunk detection combined with 2D LiDAR achieves accurate row following in orchards \cite{Shi2025}, while trunk-aware graph-SLAM and keypoint-semantic integration improve loop closure and feature distinctiveness in vineyards \cite{papadimitriou2022loop, de2025keypoint}. These studies demonstrate the value of stable semantic landmarks.

However, most fusion approaches focus on reactive navigation or optimisation-based SLAM. They do not explicitly address row aliasing within a probabilistic global localisation framework. In particular, trunk detections are typically treated as isolated landmarks rather than as structural constraints capable of disambiguating parallel corridors.


We propose a Semantic Landmark Particle Filter (SLPF) that integrates RGB-D trunk and pole detections with 2D LiDAR observations in a probabilistic formulation. The key idea is to convert stable landmarks into \emph{semantic walls}: row-level structural boundaries inferred from the spatial arrangement of trunks and support poles. Instead of treating each detection independently, neighbouring landmarks are interpreted as belonging to the same physical row, forming continuous corridor constraints that reflect the underlying vineyard layout.

These semantic walls are embedded directly within the particle-filter likelihood, strengthening discrimination between adjacent rows and explicitly penalising cross-row hypotheses under perceptual aliasing. GNSS is included as a lightweight probabilistic prior that primarily stabilises localisation in headlands where semantic observations become sparse.

The contributions of this paper are threefold. First, we introduce semantic walls, transforming sparse trunk and pole detections into row-aligned structural constraints that improve row-level discrimination. Second, we embed these constraints directly within a semantic-LiDAR particle filter, enabling explicit rejection of wrong-row hypotheses in repetitive environments. Third, we incorporate adaptive GNSS weighting based on semantic observation richness, improving robustness during headland transitions while preserving lightweight sensor requirements. Real-world experiments demonstrate improved row correctness and recovery compared to geometry-only and visual-SLAM baselines.
Code and dataset available at: \texttt{https://github.com/LCAS/Outdoor\_SLPF}.

\section{Related Work}
\label{sec:related}

Robot localisation in vineyards presents several challenges due to the environment's repetitive row structure, seasonal appearance changes, and uneven terrain conditions \cite{aguiar2022localization, nehme2021lidar}. Visual sensors such as LiDAR and cameras are prone to perceptual aliasing in repetitive environments and are sensitive to lighting variation \cite{hroob2021benchmark}. Non-visual sensors, including GNSS and ultrasonic systems, can be affected by canopy occlusion, multipath, and signal degradation \cite{agronomy11020287, costley2020landmark}. Robust localisation remains essential for enabling long-term autonomy and digital twin applications in vineyard management \cite{polvara2024bacchus, kunze2018artificial}.

LiDAR-based particle filters such as Adaptive Monte Carlo Localization (AMCL) \cite{dellaert1999icra} have been widely applied in structured environments and extended to agricultural domains \cite{astolfi2018vineyard, s22239095}. While their probabilistic formulation allows tolerance to map-sensor mismatch \cite{blok2019robot}, their measurement models rely purely on geometric consistency. In vineyards, where parallel rows generate near-identical range signatures, geometry-only likelihoods may maintain persistent wrong-row hypotheses, particularly during headland transitions.

Visual SLAM approaches have also been explored in vineyards \cite{kokas2024multicamera, papadimitriou2022loop}. Papadimitriou et al.~\cite{papadimitriou2022loop} incorporate trunk detections into a graph-SLAM formulation, demonstrating improved loop closure robustness compared to non-semantic baselines. Similarly, De Silva et al.~\cite{de2025keypoint} introduce keypoint-semantic integration to improve visual feature distinctiveness in repetitive agricultural environments. These works highlight the value of semantically stable landmarks such as vine trunks, yet they remain embedded within optimisation-based SLAM back-ends rather than explicitly reshaping the measurement likelihood in a global probabilistic filter.

Several recent works explore LiDAR-vision fusion for row navigation. For example, Shi et al.~\cite{Shi2025} combine YOLO-based trunk detection with 2D LiDAR via ray projection to extract navigation lines in dense orchard environments, achieving high lateral accuracy. Such approaches effectively support corridor following but assume locally continuous vegetation and are primarily designed for reactive control rather than maintaining a global pose posterior. Stable-point segmentation methods further distinguish permanent structural elements from seasonal vegetation to enhance long-term consistency in 3D LiDAR mapping \cite{Hroob2024}. While improving robustness to appearance changes, these approaches typically operate within dense SLAM frameworks and do not directly address row-level aliasing in global localisation.

GNSS-based localisation remains common in agricultural robotics \cite{9249176}. Sensor-fusion approaches combining SLAM and GNSS dynamically adjust weighting based on signal quality metrics \cite{Zhou2025}. While effective in GNSS-degraded scenarios, such methods often assume RTK precision or rely on factor-graph optimisation.

In contrast to prior work, our approach makes three conceptual shifts. First, rather than treating trunk detections as isolated landmarks, we connect adjacent stable landmarks into piecewise-linear semantic walls, forming row-aligned structural constraints that persist even when individual plants are missing. Second, these semantic walls are embedded directly within the particle-filter measurement likelihood, enabling explicit penalisation of cross-row hypotheses under perceptual aliasing. Third, GNSS is incorporated as an adaptive probabilistic prior whose influence depends on semantic observation richness, providing global anchoring without dominating the estimation process.

Together, these design choices specifically address the row-aliasing failure mode characteristic of repetitive agricultural environments. More broadly, they illustrate how semantically stable, spatially organised landmarks can be converted into structural constraints directly within a probabilistic measurement model. While demonstrated in vineyards, this formulation is applicable to other structured outdoor domains exhibiting parallel or repetitive geometry, such as orchards, forestry corridors, or plantation fields.

\section{Semantic Particle Filter Localisation in Vineyards}
\label{sec:main}\label{sec:method}

Vineyards exhibit strong perceptual aliasing: adjacent rows share repeated geometry, making geometry-only localisation prone to convergence to incorrect corridors when observations are sparse or ambiguous. We therefore formulate localisation as a \emph{semantic landmark particle filter} (SLPF), a recursive Bayesian estimator that maintains multiple pose hypotheses and evaluates them using motion and semantically informed measurement models.

\begin{figure}[!h]
  \centering
  \includegraphics[width=0.99\linewidth]{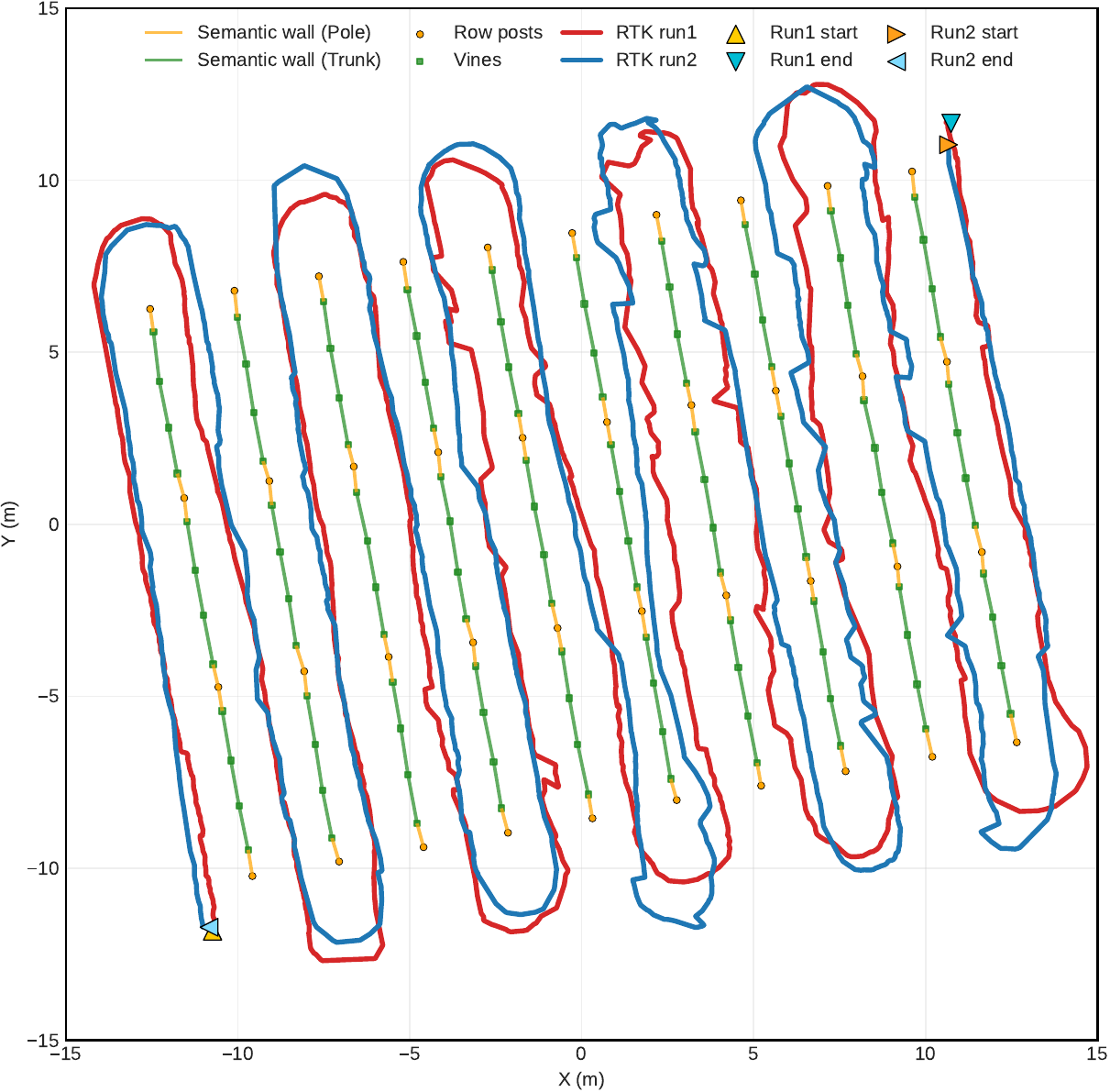}
  \caption{Vineyard structural map with RTK-GNSS ground-truth traverses for the two experiments. Orange circles and green squares indicate row posts and vines, while orange and green line segments denote pole- and trunk-based semantic walls. Red (Experiment~1) and blue (Experiment~2) trajectories are the RTK paths; start/end markers indicate traversal direction.}
  \label{fig:wallmap}
\end{figure}

\begin{figure}[t]
  \centering
  \includegraphics[width=0.98\linewidth]{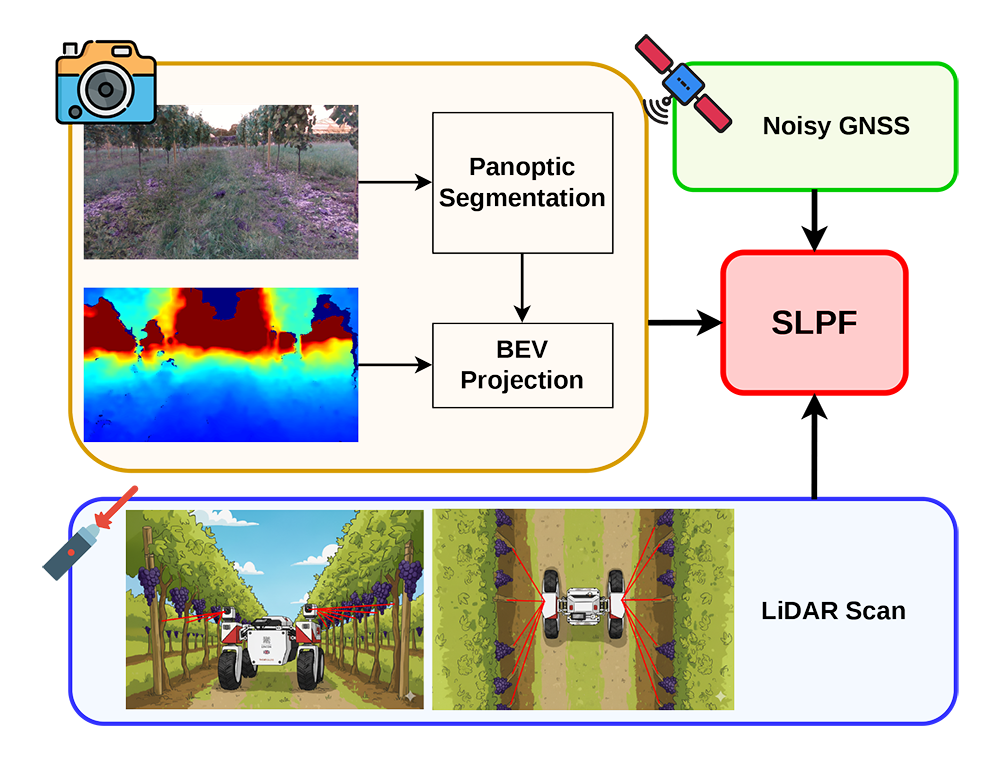}
  \caption{Overview of the Semantic Particle Filter. Instance masks from segmentation are combined with depth to generate BEV projections of vineyard landmarks. These projections label near-field LiDAR returns as semantic observations for the particle filter, augmented with a NoisyGNSS prior.}
  \label{fig:overall}
\end{figure}

Our approach leverages long-term stable vineyard landmarks, such as vine trunks and support poles, detected via instance segmentation and projected into a local bird's-eye view (BEV). These landmarks are used to semantically label near-field LiDAR returns, producing class-conditioned observations. Observations are evaluated against a surveyed row-aligned structural map derived from RTK-GNSS landmarks (Fig.~\ref{fig:wallmap}), while a NoisyGNSS prior provides a soft global anchor when semantic evidence is weak (e.g., in headlands). Fig.~\ref{fig:overall} summarises the pipeline.

\subsection{Semantic Landmark Detection and BEV Projection}

We focus on two long-term stable landmark classes: \emph{support poles} and \emph{vine trunks}. A YOLOv9-based instance segmentation model is trained on the SemanticBLT dataset~\cite{yolov9,semanticblt_dataset}. The segmentation module is modular and can be replaced without modifying the downstream particle-filter formulation.

For each detected instance \(i\) with mask \(M_i\), its range is estimated conservatively using the closest valid depth:
\begin{equation}
  d_i = \min \{ D(p)\mid p\in M_i,\; D(p)>0 \}
\end{equation}
Using camera intrinsics \(K\), a representative pixel is back-projected into the camera frame and transformed into the robot frame via known extrinsics. Ground-plane coordinates \(\mathbf{o}_i=[l_i,f_i]^\top\) (left, forward) are retained. To account for segmentation and projection uncertainty, each landmark is inflated to a circular BEV region of radius \(\mathcal{R}_{\mathrm{sem}}\).

Let \(\mathcal{P}_t=\{\mathbf{q}_j\}\) denote LiDAR points projected onto the ground plane. Each point receives a semantic label \(c_j\in\{background,pole,trunk\}\) based on BEV membership and is represented in polar form:
\begin{equation}
  r_j=\|\mathbf{o}_j\|_2,\qquad
  \phi_j=\operatorname{atan2}(l_j,f_j)
\end{equation}
Observations are grouped into semantic sets \(\mathcal{Z}^{\text{pole}}_t\), \(\mathcal{Z}^{\text{trunk}}_t\), and background \(\mathcal{Z}^{\text{bg}}_t\), with optional downsampling of background rays for efficiency.

\subsection{Row-aligned Structural Map}

Adjacent landmarks within the same row are connected into line segments, referred to as semantic walls. Unlike dense map representations, semantic walls encode only stable row boundaries (RTK-GNSS-surveyed trunks and poles) and intentionally ignore transient vegetation structure. These walls encode row continuity explicitly and provide stronger geometric discrimination between neighbouring corridors during ray casting.

Landmarks are grouped by row ID and sorted along the dominant axis; consecutive landmarks \(\{a,b\}\) define a segment \(\overline{ab}\). A segment class follows a conservative pole-dominance rule:
\begin{equation}
  class_{\overline{ab}}=
  \begin{cases}
    pole & \text{if } c_a \text{ or } c_b = pole,\\
    trunk & \text{otherwise}
  \end{cases}
\end{equation}




\subsection{Particle Filter Formulation}

\subsubsection{Motion Model}

Given odometry increments \((\Delta d_t,\Delta\theta_t)\), particle \(i\) is propagated as
\begin{equation}
  \begin{aligned}
    x_t^{(i)} &= x_{t-1}^{(i)} + \Delta d_t\cos\theta_{t-1}^{(i)} + \epsilon_x,\\
    y_t^{(i)} &= y_{t-1}^{(i)} + \Delta d_t\sin\theta_{t-1}^{(i)} + \epsilon_y,\\
    \theta_t^{(i)} &= \mathrm{wrap}(\theta_{t-1}^{(i)}+\Delta\theta_t+\epsilon_\theta)
  \end{aligned}
\end{equation}
with Gaussian noise terms.

\subsubsection{Ray Casting Prediction}

For observation \(j\) with bearing \(\phi_j\), particle \(i\) predicts
\begin{equation}
  \mathbf{d}_{ij}=
  \begin{bmatrix}
    \cos(\theta_t^{(i)}+\phi_j)\\
    \sin(\theta_t^{(i)}+\phi_j)
  \end{bmatrix}
\end{equation}
Intersecting this ray with semantic wall segments yields predicted range \(\hat r_{ij}\) and class \(\hat c_{ij}\); if no intersection occurs within \(r_{\max}\), a no-hit is recorded.

\subsection{Likelihood Estimation}

The likelihood integrates semantic rays, background free-space constraints, GNSS anchoring, and a structural corridor prior.

\subsubsection{Semantic and Background Rays}

For semantic observations:
\begin{equation}
  \psi^{\text{sem}}_{ij}=
  \begin{cases}
    -\dfrac{(r_j-\hat r_{ij})^2}{2\sigma_{\text{sem}}^2}, & \hat c_{ij}=c_j,\\
    -\dfrac{p_{\text{wrong}}^2}{2\sigma_{\text{sem}}^2}, & \hat c_{ij}\neq c_j,\hat c_{ij}\neq -1,\\
    -\dfrac{p_{\text{miss}}^2}{2\sigma_{\text{sem}}^2}, & \hat c_{ij}=-1
  \end{cases}
\end{equation}

Let \(J_t\) denote the number of retained rays at time \(t\). Aggregating over rays,
\begin{equation}
  \ell^{(i)}_{\text{sem},t}
  =
  \frac{1}{J_t}\sum_{j=1}^{J_t} w_{c_j}\,\psi_{ij}
\end{equation}
where \(w_{c_j}\) assigns relative importance to each class.

Background rays enforce free-space consistency by penalising predicted obstacles closer than observed ranges.

\subsubsection{GNSS Prior}

GNSS provides a soft global constraint:
\begin{equation}
  \ell^{(i)}_{\text{gnss},t}
  =
  -\frac{\|\mathbf{p}^{(i)}_t-\mathbf{z}^{\text{gnss}}_t\|_2^2}{2\sigma_{\text{gnss}}^2}
\end{equation}
where \(\mathbf{p}^{(i)}_t=[x_t^{(i)},y_t^{(i)}]^\top\).

\subsubsection{Corridor Prior}

Let \(d^{(i)}_{\perp,t}\) be the perpendicular distance from particle \(i\) to the nearest wall segment, and \(\Delta\theta^{(i)}_t\) the heading difference between the particle and the segment direction. The corridor term is
\begin{equation}
  \ell^{(i)}_{\text{corr},t}
  =
  -\frac{(d^{(i)}_{\perp,t})^2}{2\sigma_d^2}
  -\frac{(1-|\cos\Delta\theta^{(i)}_t|)^2}{2\sigma_h^2}
\end{equation}
This weak structural prior suppresses cross-row hypotheses while complementing sensor evidence. Moreover, it does not enforce lane following but softly regularises pose hypotheses toward feasible inter-row regions.

\subsection{Robust Fusion and Adaptive Weighting}

Each likelihood component is robustly normalised across particles:
\begin{equation}
  \tilde\ell = \mathrm{clip}\!\left(
    \frac{\ell-\mathrm{median}(\ell)}{1.4826\,\mathrm{MAD}(\ell)+\varepsilon},
  -c,c\right)
\end{equation}
where \(c>0\) bounds extreme values.

GNSS weight adapts to semantic observation richness:
\begin{equation}
  \alpha_t
  =
  \mathrm{clip}\!\left(
    \frac{1}{1+N_{\mathrm{sem},t}/K},
  \alpha_{\min},\alpha_{\max}\right)
\end{equation}
We set $\alpha_{\min} = 0.05$ and $\alpha_{\max} = 0.95$; the nominal GNSS weight of $0.5$ corresponds to the mid-range value under typical semantic observation counts.

The fused score is
\begin{equation}
  \ell^{(i)}_t
  =
  \alpha_t\,\tilde\ell^{(i)}_{\text{gnss},t}
  + (1-\alpha_t)\Big[(1-\lambda_c)\,\tilde\ell^{(i)}_{\text{sem},t}
  +\lambda_c\,\tilde\ell^{(i)}_{\text{corr},t}\Big]
\end{equation}

Particle weights are obtained via tempered softmax:
\begin{equation}
  w_t^{(i)}
  =
  \frac{\exp(\ell_t^{(i)}/\tau)}
  {\sum_k \exp(\ell_t^{(k)}/\tau)}
\end{equation}

\subsection{Resampling and pose estimation}

Effective sample size is computed as
\begin{equation}
  N_{\text{eff}} = \frac{1}{\sum_i (w_t^{(i)})^2}
\end{equation}
Resampling is triggered when \(N_{\text{eff}}\) falls below a threshold; KLD sampling determines the required number of particles. The pose estimate is computed via weighted mean for position and circular mean for heading, reverting to the MAP particle when circular variance is high.

\subsection{Computational Considerations}

The dominant cost is ray–segment intersection, scaling as \(\mathcal{O}(N_p\,J_t\,N_s)\). Background rays are downsampled and batched intersection with segment chunking is used to maintain tractable runtime.

\section{Experimental Evaluation}
\label{sec:exp}


We evaluate SLPF in a controlled vineyard environment to assess its ability to resolve row-level perceptual aliasing under realistic sensing conditions.

\subsection{Experimental Setup}
Experiments were conducted using a Thorvald robot platform (Saga Robotics) equipped with a front-mounted Intel RealSense D435i RGB-D camera, a 2D LiDAR (SICK Tim7), and an RTK-GNSS receiver. The robot traversed a compact test vineyard consisting of 10 rows, each approximately 16.86\,m long. The vineyard covers about 385\,m$^2$ (0.0385\,ha), corresponding to an approximate footprint of 17.0\,m $\times$ 22.6\,m with a mean row spacing of $\sim$2.50\,m. During the trials, synchronised RGB-D images, LiDAR scans, and RTK ground-truth poses were recorded.

To emulate realistic noisy GNSS conditions, the RTK trajectory was degraded using a stochastic noise model capturing three common error modes: (i) slowly varying drift (temporally correlated bias), (ii) short-term receiver noise (zero-mean Gaussian), and (iii) occasional jump-like multipath errors (heavy-tailed outliers)~\cite{groves2013principles,kaplan2017understanding,brown2012introduction}. We report results over three independent noise realisations while preserving original timestamps and orientations.

Evaluation is performed on two traversals using the same semantic-wall map (see Fig.~\ref{fig:wallmap}). For stochastic components, three random seeds (\{11,22,33\}) are used and statistics are aggregated across seeds; representative trajectory plots use seed~11. We compare SLPF against NoisyGNSS, AMCL, AMCL+NoisyGNSS (Kalman fusion), and RTAB-Map (RGB and RGBD)~\cite{rtabmap}. AMCL is run with default parameters, except that LiDAR maximum range is truncated to \(5\)~m to match the semantic sensing range used by SLPF, ensuring comparable observation horizons. RTAB-Map (RGB/RGBD) uses default configuration files.

SLPF parameters are kept fixed across both traversals to avoid environment-specific tuning. Key settings include miss/wrong-hit penalties (4.0/4.0), GNSS weight 0.5 with adaptive scaling \(K=4\), semantic noise \(\sigma_{\mathrm{sem}}=0.05\), GNSS noise \(\sigma_{\mathrm{gnss}}=1.1\)~m, corridor weight 0.30, background weight 0.20, particle count \(N_p=100\), frame stride 4, and maximum semantic range \(r_{\max}=5\)~m. Pose smoothing and yaw filtering parameters are fixed across runs, and CUDA acceleration is enabled without visualisation.

Performance is evaluated using raw and aligned Absolute Pose Error (APE), Relative Pose Error (RPE at 2\,m and 5\,m segments), cross-track error (XT), row-correct fraction, and row mislocalisation events.




\subsection{Results and Discussion}
\label{sec:results}

\begin{figure*}[!ht]
  \centering
  \includegraphics[width=0.99\linewidth]{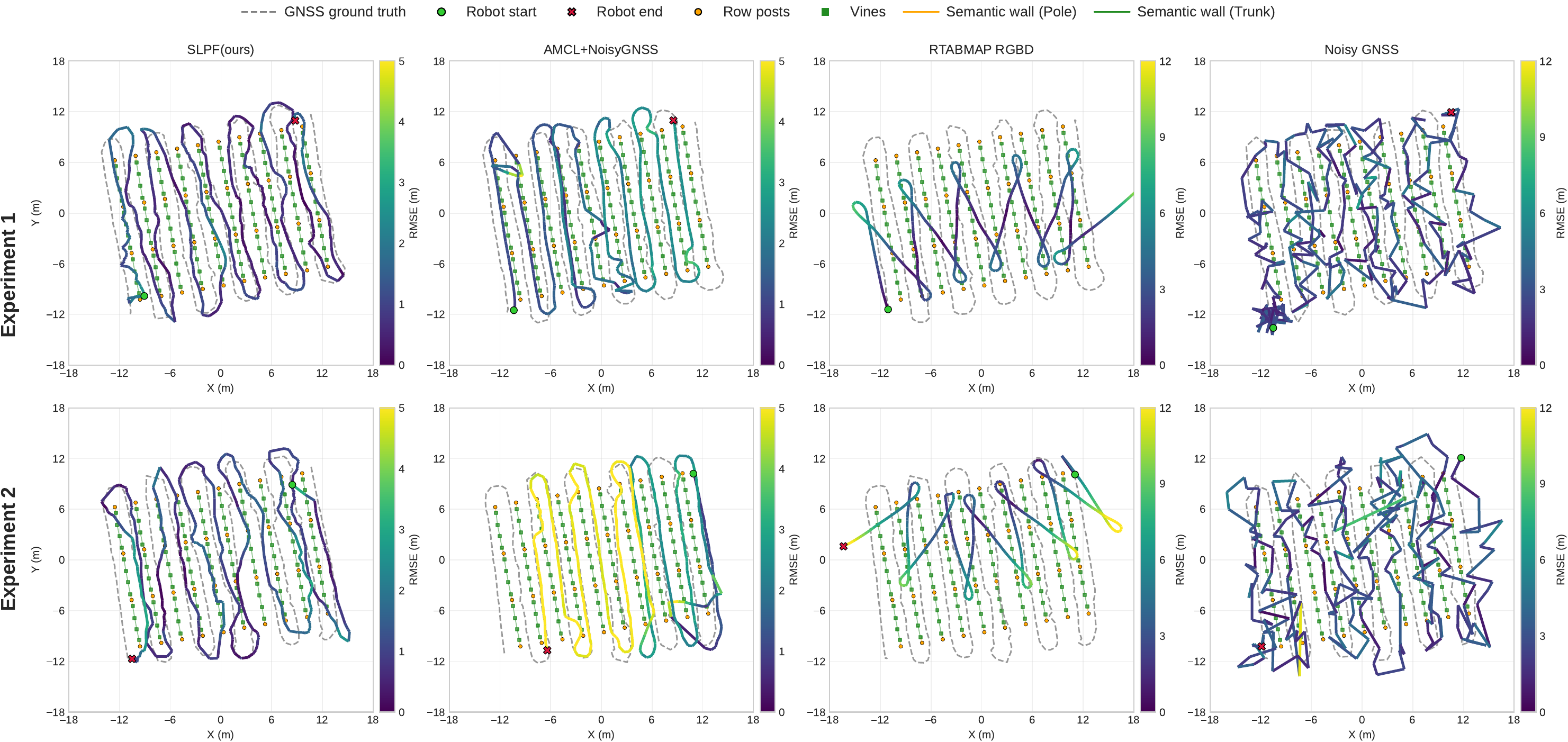}
  \caption{Raw Absolute Pose Error (APE) trajectories comparing SLPF (ours), AMCL+NoisyGNSS (Kalman fusion), RTAB-Map RGBD, and NoisyGNSS (subsampled to facilitate visualisation). The dashed black line is ground truth; coloured paths show method deviations.}
  \label{fig:plots}
\end{figure*}

\begin{table*}[t]
  \centering
  \caption{Localisation performance across methods using absolute/relative pose errors (APE/RPE), cross-track error, and row-consistency metrics. Values are reported as mean$\pm$std over three runs. 
  }
  \setlength{\tabcolsep}{4.5pt}
  \begin{adjustbox}{max width=\textwidth}
    \begin{tabular}{llcccc ccc cc}
      \toprule
      \multirow{2}{*}{} &
      \multirow{2}{*}{\textbf{Method}} &
      \multicolumn{2}{c}{\textbf{APE RMSE$\downarrow$ [m]}} &
      \multicolumn{2}{c}{\textbf{RPE RMSE$\downarrow$ [m]}} &
      \multicolumn{3}{c}{\textbf{Cross-track$\downarrow$ [m]}} &
      \textbf{Row correct$\uparrow$} &
      \textbf{Row mislocalisation$\downarrow$}\\
      \cmidrule(lr){3-4}\cmidrule(lr){5-6}\cmidrule(lr){7-9}
      & & Raw & Aligned & 2\,m & 5\,m & Mean & Median & Max & Fraction & \# events\\
      \cmidrule(lr){2-11}
      \multirow{7}{*}{\rotatebox{90}{\textbf{Experiment 1}}} & NoisyGNSS & 3.04$\pm$0.13 & 2.98$\pm$0.13 & 3.56$\pm$0.29 & 3.99$\pm$0.43 & 1.77$\pm$0.12 & 1.78 & 1.78 & 0.64$\pm$0.04 & 64.3$\pm$5.4\\
      & AMCL & 1.37$\pm$0.47 & 1.02$\pm$0.30 & 4.69$\pm$0.05 & 5.39$\pm$0.11 & 1.40$\pm$0.10 & 0.84 & 1.54 & 0.67$\pm$0.13 & 27.3$\pm$3.1\\
      & AMCL+NoisyGNSS & 1.33$\pm$0.46 & 0.99$\pm$0.30 & 4.66$\pm$0.04 & 5.35$\pm$0.10 & 1.39$\pm$0.09 & \textbf{0.80} & \textbf{1.52} & 0.67$\pm$0.13 & 27.3$\pm$3.1\\
      & RGB RTAB-Map & 59.62$\pm$0.48 & 6.68$\pm$0.03 & \textbf{1.19$\pm$0.01} & \textbf{2.21$\pm$0.01} & 6.21$\pm$0.03 & 6.15 & 19.49 & 0.45$\pm$0.00 & 14.0$\pm$0.0\\
      & RGBD RTAB-Map & 61.33$\pm$0.28 & 10.02$\pm$0.03 & 8.25$\pm$0.08 & 10.79$\pm$0.11 & 6.78$\pm$0.01 & 6.13 & 46.10 & 0.48$\pm$0.00 & \textbf{13.3$\pm$0.5}\\
      \cmidrule(lr){2-11}
      & SLPF (ours) & \textbf{1.07$\pm$0.09} & \textbf{1.04$\pm$0.10} & 3.33$\pm$0.07 & 6.92$\pm$0.09 & \textbf{1.26$\pm$0.06} & 1.25 & 3.85 & \textbf{0.73$\pm$0.01} & 34.67$\pm$1.70\\
      \cmidrule(lr){1-11}
      \multirow{7}{*}{\rotatebox{90}{\textbf{Experiment 2}}} & NoisyGNSS & 3.16$\pm$0.22 & 3.09$\pm$0.23 & 3.76$\pm$0.32 & 4.22$\pm$0.55 & 1.99$\pm$0.04 & 1.80 & 36.99 & 0.58$\pm$0.02 & 713.3$\pm$23.2\\
      & AMCL & 3.50$\pm$0.94 & 2.04$\pm$0.31 & 2.78$\pm$0.79 & 3.43$\pm$0.92 & 1.55$\pm$0.12 & 1.86 & 7.01 & 0.55$\pm$0.05 & 26.7$\pm$2.1\\
      & AMCL+NoisyGNSS & 3.38$\pm$0.91 & 1.98$\pm$0.31 & 2.80$\pm$0.78 & 3.42$\pm$0.92 & 1.51$\pm$0.11 & 1.77 & 6.38 & 0.55$\pm$0.05 & 26.7$\pm$1.2\\
      & RGB RTAB-Map & 85.95$\pm$0.19 & 9.12$\pm$0.41 & \textbf{1.63$\pm$0.01} & \textbf{2.48$\pm$0.00} & 6.81$\pm$0.31 & 6.26 & 20.80 & 0.39$\pm$0.02 & \textbf{17.0$\pm$0.8}\\
      & RGBD RTAB-Map & 87.17$\pm$0.01 & 9.06$\pm$0.00 & 1.81$\pm$0.00 & 3.43$\pm$0.00 & 7.25$\pm$0.08 & 6.73 & 38.15 & 0.42$\pm$0.00 & 18.7$\pm$0.5\\
      \cmidrule(lr){2-11}
      & SLPF (ours) & \textbf{1.24$\pm$0.04} & \textbf{1.11$\pm$0.06} & 3.34$\pm$0.02 & 6.82$\pm$0.07 & \textbf{1.46$\pm$0.03} & \textbf{1.31} & \textbf{4.35} & \textbf{0.67$\pm$0.02} & 28.0$\pm$3.3\\
      \bottomrule
    \end{tabular}
  \end{adjustbox}
  \label{tab:evo}
\end{table*}

Across both traversal directions, SLPF consistently achieves the lowest raw APE among all methods. In Experiment~1, SLPF attains \(1.07\pm0.09\)~m, improving over AMCL (\(1.37\pm0.47\)~m) and AMCL+NoisyGNSS (\(1.33\pm0.46\)~m), and substantially outperforming NoisyGNSS (\(3.04\pm0.13\)~m). In Experiment~2, SLPF remains stable at \(1.24\pm0.04\)~m, whereas AMCL and AMCL+NoisyGNSS degrade to \(3.50\pm0.94\)~m and \(3.38\pm0.91\)~m, respectively. Overall, SLPF reduces APE by 22\% (Experiment~1) and 65\% (Experiment~2) relative to AMCL, and by 65\% and 61\% relative to NoisyGNSS.

These improvements translate directly to row-level metrics. SLPF attains the highest row-correct fraction in both experiments (\(0.73\pm0.01\) and \(0.67\pm0.02\)), exceeding AMCL-based baselines and RTAB-Map variants. Mean cross-track error is also lowest for SLPF (\(1.26\pm0.06\)~m and \(1.46\pm0.03\)~m), indicating tighter adherence to the correct corridor.

Fig.~\ref{fig:plots} illustrates the qualitative behaviour. AMCL and AMCL+NoisyGNSS produce smooth trajectories but remain vulnerable to geometric aliasing: once drifting into an adjacent row, recovery is limited. SLPF mitigates this failure mode through semantic wall constraints and adaptive GNSS weighting, enabling recovery from wrong-row hypotheses. NoisyGNSS remains globally bounded but lacks structural information, resulting in lower row correctness. RTAB-Map achieves low short-horizon RPE, consistent with locally smooth tracking; however, its large raw APE and reduced row-correct fraction indicate global row misalignment, particularly during headland transitions where visual similarity is amplified. Using SE(3)-aligned APE, the RMSE decreases to approximately 10\,m, confirming locally consistent tracking but persistent global row misalignment in the surveyed map frame. Default RTAB-Map configurations are retained to reflect typical deployment rather than environment-specific tuning.

Row-mislocalisation events should be interpreted jointly with row-correct and cross-track metrics, as they count transitions rather than duration. Although AMCL-based methods report fewer switch events in Experiment~1, SLPF achieves higher row correctness and lower cross-track error, indicating shorter and recoverable wrong-row episodes. Importantly, ablation results (Table~\ref{tab:spfpp_ablation}) show that removing the corridor term degrades performance but does not eliminate the gains from semantic likelihood integration, indicating that the primary improvement arises from structural semantic modelling rather than corridor regularisation alone.

\paragraph{Failure Mode Analysis}

\begin{figure}[t]
    \centering
    \includegraphics[width=0.95\linewidth]{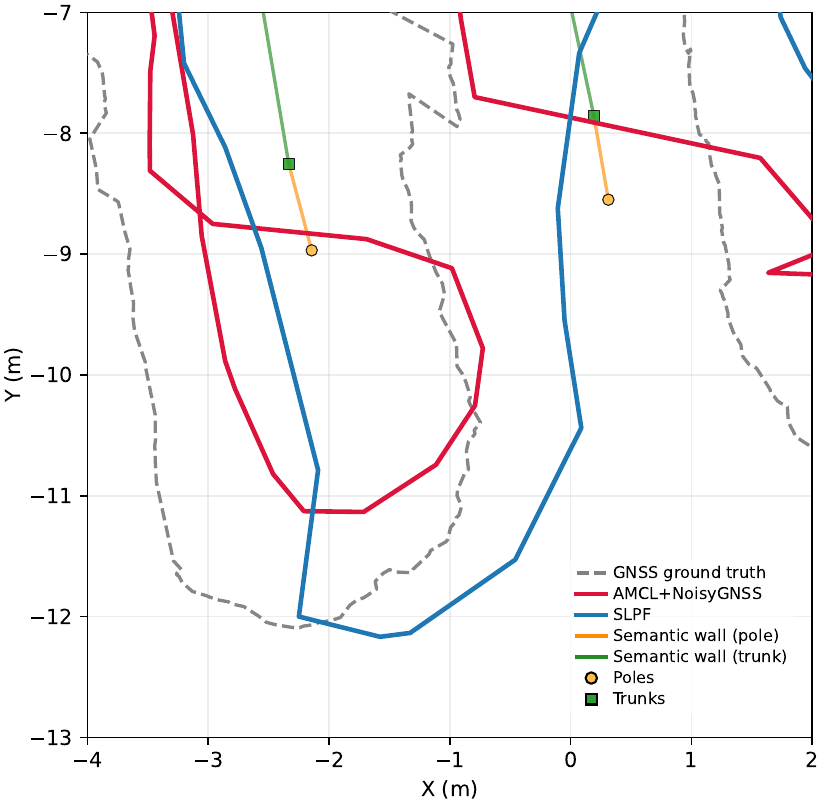}
    \caption{Zoomed comparison during a headland transition (Experiment~1). AMCL+NoisyGNSS remains aligned to an adjacent row despite locally smooth tracking, while SLPF recovers through semantic wall constraints and adaptive GNSS weighting.}
    \label{fig:headland_zoom}
\end{figure}

Row-level aliasing is most pronounced during headland transitions, where rapid heading changes and reduced landmark visibility increase ambiguity between adjacent corridors. In these conditions, geometry-only filters may converge to a neighbouring row that produces locally consistent LiDAR ranges. Because parallel rows yield similar ray signatures, such hypotheses can persist, resulting in smooth trajectories with low short-horizon RPE but large raw APE and reduced row correctness (see Fig.~\ref{fig:headland_zoom}).
SLPF mitigates this behaviour through structural discrimination. When semantic observations reappear, the wall-based likelihood penalises cross-row inconsistencies, causing incorrect particle clusters to collapse. The adaptive GNSS prior stabilises pose during low-semantic intervals without overwhelming corridor constraints.

These results indicate that SLPF directly addresses the aliasing-induced wrong-row failure mode rather than merely reducing average localisation error.

\paragraph{Performance Analysis}On our test workstation (AMD Ryzen Threadripper 2950X CPU, 64\,GB RAM, NVIDIA RTX 4060 Ti / RTX 3070 GPU); the full SLPF pipeline runs at \(13.04\pm0.21\) Hz (stride-adjusted: \(52.18\pm0.84\) Hz). Semantic inference dominates runtime (44.35 ms/frame), while particle-filter likelihood evaluation and resampling require approximately 8 ms/frame, supporting real-time execution in the tested configuration, with particle-filter evaluation accounting for a small fraction of total runtime.

\paragraph{Summary}Overall, Table~\ref{tab:evo} and Fig.~\ref{fig:plots} demonstrate that SLPF provides improved global consistency and corridor adherence across traversal directions. These results suggest that the proposed semantic-likelihood formulation is particularly beneficial under increased perceptual ambiguity, rather than being specific to a given traversal configuration.

\subsection{Robustness to Detection and Map Degradation}
\label{sec:robustness_run1}
We evaluate two stress tests on Experiment~1: \emph{(A)} random semantic-detection dropping
(20\%, 40\%), referred to as Detection Dropout (DetDrop), and \emph{(B)}
landmark removal from one contiguous map section (30\%, 50\%), referred to as
Section Landmark Removal (MapRemove).
Table~\ref{tab:robustness_run1} follows the same metrics/protocol used in the main comparison and reports
mean$\pm$std over three seeds.

\begin{table}[t]
  \centering
  \caption{Experiment~1 robustness under Detection Dropout and Section Landmark Removal. Values are mean$\pm$std over three runs.}
  \scriptsize
  \setlength{\tabcolsep}{3.0pt}
  \begin{tabular}{lccccc}
    \toprule
    \textbf{Variant} &
    \textbf{APE$\downarrow$} &
    \textbf{RPE 2m$\downarrow$} &
    \textbf{XT mean$\downarrow$} &
    \textbf{Row corr.$\uparrow$} &
    \textbf{Switches$\downarrow$}\\
    \midrule
    SLPF (full map) & \textbf{1.00$\pm$0.03} & \textbf{3.33$\pm$0.02} & 1.31$\pm$0.05 & 0.72$\pm$0.02 & 33.00$\pm$0.82\\
    \cmidrule(lr){1-6}
    DetDrop 20\% & 1.02$\pm$0.08 & 3.37$\pm$0.06 & 1.31$\pm$0.05 & 0.73$\pm$0.01 & 33.33$\pm$4.11\\
    DetDrop 40\% & 1.05$\pm$0.09 & 3.33$\pm$0.04 & \textbf{1.26$\pm$0.05} & \textbf{0.74$\pm$0.03} & \textbf{30.67$\pm$0.94}\\
    \cmidrule(lr){1-6}
    MapRemove 30\% & 1.05$\pm$0.04 & 3.37$\pm$0.03 & 1.27$\pm$0.02 & 0.72$\pm$0.01 & 30.33$\pm$2.05\\
    MapRemove 50\% & 1.13$\pm$0.13 & 3.36$\pm$0.06 & 1.32$\pm$0.02 & 0.72$\pm$0.01 & 31.00$\pm$0.82\\
    \bottomrule
  \end{tabular}
  \label{tab:robustness_run1}
\end{table}

For Detection Dropout (DetDrop), performance remains stable under increasing semantic sparsity. APE increases modestly from $1.00\pm0.03$\,m (baseline) to $1.02\pm0.08$\,m (20\%) and $1.05\pm0.09$\,m (40\%), while row correctness remains comparable or slightly improves ($0.72\pm0.02$ to $0.73\pm0.01$ and $0.74\pm0.03$). The number of row mislocalisation events remains within variance, indicating that moderate landmark loss does not destabilise the filter.

For Section Landmark Removal (MapRemove), performance degrades progressively with increasing map sparsity. Global APE rises from $1.00\pm0.03$\,m to $1.05\pm0.04$\,m (30\%) and $1.13\pm0.13$\,m (50\%), while row correctness remains stable at $0.72\pm0.01$. Although landmark removal locally perturbs estimation within the affected section, the system consistently recovers once re-entering mapped regions, with recovery observed in all runs and an average recovery distance of approximately 12\,m.

\subsection{Ablation Study}




\begin{table*}[t]
  \centering
  \small
  \caption{SLPF ablation study (mean$\pm$std across 3 seeds).
  }
  \setlength{\tabcolsep}{4.5pt}
  \begin{adjustbox}{max width=\textwidth}
    \begin{tabular}{llcccc ccc}
      \toprule
      \multirow{2}{*}{} &
      \multirow{2}{*}{\textbf{Variant}} &
      \multicolumn{2}{c}{\textbf{APE RMSE$\downarrow$ [m]}} &
      \multicolumn{2}{c}{\textbf{RPE RMSE$\downarrow$ [m]}} &
      \multicolumn{1}{c}{\textbf{Cross-track$\downarrow$ [m]}} &
      \textbf{Row correct$\uparrow$} &
      \textbf{Row mislocalisation$\downarrow$}\\
      \cmidrule(lr){3-4}\cmidrule(lr){5-6}\cmidrule(lr){7-7}
      & & Raw & Align & 2\,m & 5\,m & Mean & Fraction & \# events\\
      \cmidrule(lr){2-9}
      \multirow{10}{*}{\rotatebox{90}{\textbf{Experiment 1}}} & non\_wall\_points & \textbf{0.98$\pm$0.08} & \textbf{0.95$\pm$0.07} & 3.34$\pm$0.07 & 6.91$\pm$0.06 & 1.30$\pm$0.03 & 0.70$\pm$0.03 & 32.67$\pm$2.49 \\
      & static\_gnss\_weight & 1.16$\pm$0.12 & 1.13$\pm$0.11 & 3.38$\pm$0.06 & 6.94$\pm$0.17 & 1.26$\pm$0.04 & \textbf{0.74$\pm$0.03} & 31.33$\pm$1.70 \\
      & no\_pose\_smoothing & 1.11$\pm$0.01 & 1.08$\pm$0.00 & \textbf{3.19$\pm$0.03} & \textbf{6.55$\pm$0.09} & \textbf{1.24$\pm$0.00} & 0.71$\pm$0.01 & 40.67$\pm$0.47 \\
      & poles\_only & 1.15$\pm$0.20 & 1.15$\pm$0.20 & 3.37$\pm$0.06 & 7.00$\pm$0.21 & 1.29$\pm$0.06 & 0.69$\pm$0.03 & 32.33$\pm$1.25 \\
      & trunks\_only & 1.04$\pm$0.14 & 1.04$\pm$0.14 & 3.35$\pm$0.10 & 6.96$\pm$0.11 & 1.29$\pm$0.05 & 0.72$\pm$0.02 & 33.33$\pm$1.25 \\
      & no\_background & 1.44$\pm$0.53 & 1.31$\pm$0.42 & 3.30$\pm$0.05 & 6.63$\pm$0.11 & 1.43$\pm$0.06 & 0.67$\pm$0.03 & 33.67$\pm$1.25 \\
      & no\_corridor & 1.37$\pm$0.12 & 1.34$\pm$0.12 & 3.28$\pm$0.03 & 6.81$\pm$0.13 & 1.44$\pm$0.02 & 0.67$\pm$0.02 & 35.33$\pm$2.05 \\
      & no\_semantic & 1.64$\pm$0.64 & 1.44$\pm$0.48 & 3.31$\pm$0.11 & 6.91$\pm$0.01 & 1.50$\pm$0.20 & 0.67$\pm$0.02 & 33.00$\pm$0.00 \\
      & no\_gnss & 7.55$\pm$4.87 & 4.28$\pm$1.03 & 3.26$\pm$0.11 & 6.65$\pm$0.32 & 3.18$\pm$0.84 & 0.32$\pm$0.08 & \textbf{20.00$\pm$5.10} \\
      \cmidrule(lr){2-9}
      & \textbf{full (SLPF)} & 1.07$\pm$0.09 & 1.04$\pm$0.10 & 3.33$\pm$0.07 & 6.92$\pm$0.09 & 1.26$\pm$0.06 & 0.73$\pm$0.01 & 34.67$\pm$1.70 \\
      \bottomrule
    \end{tabular}
  \end{adjustbox}
  \label{tab:spfpp_ablation}
\end{table*}

Table~\ref{tab:spfpp_ablation} reports the principal ablations of SLPF on Experiment~1. The full configuration provides the most balanced performance across global accuracy, corridor adherence, and row consistency.

\paragraph{Semantic Walls vs.\ Point-based Matching}
The \texttt{non\_wall\_points} variant replaces semantic walls with point-to-point semantic matching. While it attains slightly lower raw APE, row correctness decreases relative to the full method. This indicates that local geometric fitting alone is insufficient under perceptual aliasing, whereas encoding row continuity through semantic walls improves discrimination between adjacent corridors.

\paragraph{Poles-only vs.\ Trunks-only Semantics}
The \texttt{poles\_only} and \texttt{trunks\_only} variants isolate each semantic class. Trunks-only achieves higher row correctness (0.72) than poles-only (0.69), suggesting that trunks provide denser and more consistently observable row-aligned cues. However, both underperform the full configuration, demonstrating that combining semantic classes yields a more stable likelihood across varying visibility conditions.

\paragraph{Role of the GNSS Prior}
Removing GNSS (\texttt{no\_gnss}) leads to substantial degradation in APE, cross-track error, and row correctness, confirming that the GNSS prior is critical for maintaining global consistency when semantic evidence weakens (e.g., headlands). The reduced number of row-switch events reflects prolonged persistence in an incorrect row rather than improved stability. Using a fixed GNSS weight (\texttt{static\_gnss\_weight}) slightly increases row correctness but worsens APE, supporting the use of dynamic weighting to balance local semantic evidence with global priors.

\paragraph{Pose Smoothing}
Disabling pose smoothing (\texttt{no\_pose\_smoothing}) marginally improves short-horizon RPE but increases row mislocalisation events. This suggests that smoothing suppresses transient row-to-row oscillations even if it slightly penalises short-term relative accuracy.

\paragraph{Background and Corridor Terms}
Removing the background or corridor terms (\texttt{no\_background}, \texttt{no\_corridor}) degrades APE, cross-track error, and row correctness. These components therefore act as structural regularisers: background consistency constrains free-space interpretation, while the corridor term biases particles toward feasible inter-row regions, improving robustness under repetitive geometry.

\paragraph{Semantic Likelihood}
Finally, removing semantic likelihood cues entirely (\texttt{no\_semantic}) degrades all metrics relative to the full system, demonstrating that semantic information is a primary contributor to SLPF performance beyond the GNSS prior alone.


\subsection{Limitations}

The proposed formulation assumes the availability of a surveyed semantic-wall map and reliable detection of stable landmarks. Severe occlusion, structural reconfiguration, or prolonged degradation of both semantic and GNSS signals may reduce global consistency. While robustness experiments demonstrate graceful degradation, future work will investigate dynamic map updating and cross-season adaptation.

\section{Conclusion}
\label{sec:conclusion}

We presented a Semantic Landmark Particle Filter (SLPF) for robust localisation in vineyards affected by row-level perceptual aliasing. By detecting stable landmarks (trunks and poles) and organising them into row-aligned structural constraints, the method embeds corridor topology directly into the particle-filter likelihood, enabling explicit rejection of cross-row hypotheses in repetitive environments. A lightweight GNSS prior further stabilises localisation during headland transitions where semantic cues are sparse.

Across two traversal directions, SLPF consistently improves global map-frame accuracy and corridor adherence relative to geometry-only and GNSS-based baselines. Compared to AMCL, raw Absolute Pose Error is reduced by 22\% and 65\%; relative to a NoisyGNSS baseline, reductions of 65\% and 61\% are observed. Improvements in row correctness and cross-track error confirm more reliable recovery from wrong-row hypotheses while maintaining stable row tracking.

Future work will investigate cross-seasonal robustness through multi-season trials and domain adaptation for the panoptic segmentation model. We also aim to extend the formulation to other structured agricultural environments, such as orchards and forestry, where semantically organised landmark constraints may similarly enhance long-term localisation.

\section*{Acknowledgements}
Generative AI tools and technologies (including ChatGPT, Gemini and Claude) were used during the preparation of this manuscript to assist with (i) generating and refining visualisation assets, (ii) editing and improving the clarity of manuscript text, and (iii) supporting code development. All AI-assisted content was reviewed, edited, and verified by the authors, who take full responsibility for the final manuscript and any accompanying artefacts.

\balance
\bibliographystyle{ieeetr}

\bibliography{glorified,new}

\end{document}